\documentclass{vgtc}                          

\graphicspath{{figures/}{pictures/}{images/}{./}} 

\usepackage{times}                     

\usepackage{tabu}                      
\usepackage{booktabs}                  
\usepackage{lipsum}                    
\usepackage{mwe}                       

\usepackage{mathptmx}                  
\usepackage{amsmath}
\usepackage{amssymb}

\onlineid{1104}

\vgtccategory{Research}

\vgtcinsertpkg

\title{HGS: Hybrid Gaussian Splatting with Static-Dynamic Decomposition for Compact Dynamic View Synthesis}

\author{Kaizhe Zhang\thanks{e-mail: zkz1081@stu.xjtu.edu.cn}\\ %
        \scriptsize Xi'an Jiaotong University %
\and Yijie Zhou\\ %
        \scriptsize Xi'an Jiaotong University %
\and Weizhan Zhang\\ %
        \scriptsize Xi'an Jiaotong University %
\and Xuanyu Wang\\ %
        \scriptsize Xi'an Jiaotong University %
\and Caixia Yan\\ %
        \scriptsize Xi'an Jiaotong University %
\and Haipeng Du\\ %
        \scriptsize Xi'an Jiaotong University %
\and yugui xie\\ %
        \scriptsize chinamobile.com %
\and Yu-Hui Wen\\ %
        \scriptsize Beijing Jiaotong University %
\and Yong-Jin Liu\\ %
        \scriptsize Tsinghua University %
}

\teaser{
  \centering
  \includegraphics[width=\textwidth]{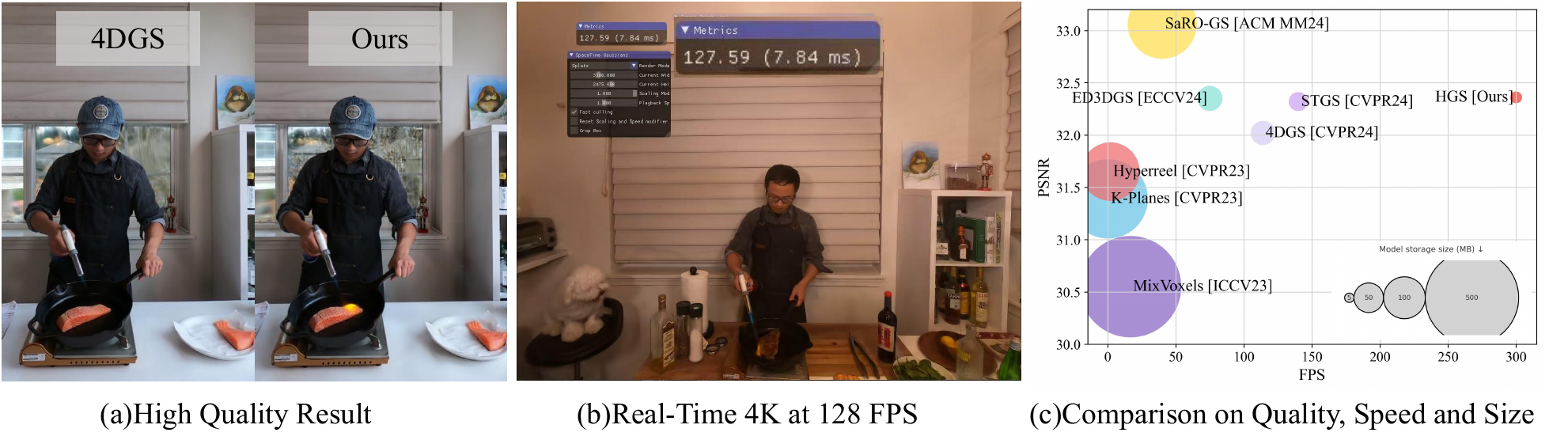}
  \caption{Performance comparison with existing SOTA~\cite{li2024spacetime, fridovich2023k, wang2023mixed, attal2023hyperreel, wu20244d, bae2024per, yan20244d}. Our approach achieves higher-quality reconstruction
in (a) temporally complex scenes, (b) while maintaining real-time rendering, (c) with an improvement in performance.}
  \label{fig:teaser}
}

\abstract{
    Dynamic novel view synthesis (NVS) is essential for creating immersive experiences. Existing approaches have advanced dynamic NVS by introducing 3D Gaussian Splatting (3DGS) with implicit deformation fields or indiscriminately assigned time-varying parameters, surpassing NeRF-based methods. However, due to excessive model complexity and parameter redundancy, they incur large model sizes and slow rendering speeds, making them inefficient for real-time applications, particularly on resource-constrained devices. To obtain a more efficient model with fewer redundant parameters, in this paper, we propose Hybrid Gaussian Splatting (HGS), a compact and efficient framework explicitly designed to disentangle static and dynamic regions of a scene within a unified representation. The core innovation of HGS lies in our Static–Dynamic Decomposition (SDD) strategy, which leverages Radial Basis Function (RBF) modeling for Gaussian primitives. Specifically, for dynamic regions, we employ time-dependent RBFs to effectively capture temporal variations and handle abrupt scene changes, while for static regions, we reduce redundancy by sharing temporally invariant parameters. Additionally, we introduce a two-stage training strategy tailored for explicit models to enhance temporal coherence at static-dynamic boundaries. Experimental results demonstrate that our method reduces model size by up to 98\% and achieves real-time rendering at up to 125 FPS at 4K resolution on a single RTX 3090 GPU. It further sustains 160 FPS at $1352{\times}1014$ on an RTX 3050 and has been integrated into the VR system. Moreover, HGS achieves comparable rendering quality to state-of-the-art methods while providing significantly improved visual fidelity for high-frequency details and abrupt scene changes.
} 


\begin{document}


\firstsection{Introduction}

\maketitle

Dynamic novel view synthesis (NVS) aims to reconstruct photorealistic images of dynamic scenes from novel viewpoints. This task has been consistently pursued in computer vision and graphics due to its broad applications in augmented reality, virtual reality, telepresence, and immersive media streaming.

Recent advances in implicit neural representations, particularly Neural Radiance Fields (NeRF)~\cite{mildenhall2021nerf}, have significantly enhanced the fidelity and convenience of static scene modeling. However, NeRF-based methods typically suffer from prolonged training times and slow rendering speed. Although substantial research~\cite{fridovich2023k, barron2021mip, liu2020neural, chen2022tensorf, chen2023neurbf, muller2022instant, fridovich2022plenoxels, sun2022direct} has aimed to mitigate these limitations, it remains challenging to achieve both high-quality and fast rendering simultaneously.

To address NeRF's computational bottlenecks from dense volume sampling, explicit approaches like 3D Gaussian Splatting (3DGS)~\cite{kerbl20233d} have emerged. 3DGS explicitly models scenes using differentiable 3D Gaussian primitives that can be efficiently rasterized onto image planes. This rasterization strategy significantly improves rendering speed while maintaining high visual fidelity, addressing NeRF's limitations in computational efficiency.
Despite the success of 3DGS in static scene modeling, its extension to dynamic scenarios remains challenging. Prior works~\cite{lee2024compact, lu2024scaffold, morgenstern2024compact} demonstrate effectiveness in static settings but encounter inter-frame discontinuities in dynamic scenes. Recent methods adopt implicit representations~\cite{huang2024sc, kratimenos2024dynmf, liang2023gaufre, wu20244d, yang2024deformable, yan20244d} to capture dynamics, but these incur high computational costs, slow rendering, and failure under abrupt motion changes. To improve efficiency, explicit techniques—such as time-dependent modules~\cite{lin2024gaussian, li2024spacetime, katsumata2023efficient} or 4D Gaussian extensions~\cite{duan20244d, yang2023real}—have been explored. However, they assign dynamic parameters to static regions, causing parameter redundancy, excessive complexity, and reduced efficiency.

In summary, existing dynamic 3DGS approaches face two key challenges: (1) excessive temporal parameters assigned to static content, resulting in significant computational and memory overhead; and (2) visual degradation due to temporal artifacts and the loss of high-frequency details in static regions. Moreover, commonly used temporal modeling is inadequate for capturing abrupt scene transitions, limiting the fidelity and robustness of dynamic scene reconstruction.

To address the aforementioned limitations, we introduce a novel Hybrid Gaussian Splatting (HGS) framework for dynamic scene representation with lower storage cost, higher rendering speed, and high quality, specifically designed for dynamic scene representation. To reduce storage cost, we adopt a Static-Dynamic Decomposition (SDD) strategy by deploying separate sets of static and dynamic Gaussian primitives, enabling explicit separation of static and dynamic regions. For static regions, we share the parameters to enable modeling without the burden of unnecessary dynamic parameters, reducing computational overhead. For dynamic regions, we restrict temporal Radial Basis Function (RBF)~\cite{li2024spacetime} to them, avoiding implicit motion modeling such as deformation fields. This scheme further reduces computational overhead and enables even faster rendering speed. 

To suppress temporal artifacts, our proposed SDD strategy constrains temporal RBF~\cite{li2024spacetime} to dynamic regions only. This constraint effectively reduces cross-region temporal interference, preserving high-frequency details and visual fidelity within static content. Our explicit separation, which disentangles dynamic motion from static regions, not only enhances motion accuracy but also significantly improves the overall rendering quality. To further strengthen consistency at the boundary between static and dynamic regions, we introduce a two-stage training strategy specifically tailored for our explicit Gaussian representation. Within each optimization cycle, the training alternates between optimizing static and dynamic primitives. This iterative optimization strategy facilitates mutual adaptation between static and dynamic components.  Combining it with modeling dynamic regions explicitly using temporal RBF, our approach effectively copes with abrupt scene changes and substantially reduces artifacts along static-dynamic boundaries.

Our HGS achieves over 300 FPS at 1080p resolution on a single RTX 3090, while requiring significantly fewer parameters, thanks to the explicit SDD strategy. The temporal RBF used for dynamic regions and our two-stage training strategy effectively reduces temporal artifacts. Therefore, with a reduced number of parameters,  our HGS achieves superior or comparable performance to recent  state-of-the-art methods in terms of both PSNR and SSIM~\cite{zhang2018unreasonable}. In most scenarios, improved visual quality can be observed, attributed to the suppression of temporal artifacts and the preservation of high-frequency details in static regions. Our main contributions are summarized as follows:
\begin{itemize}
    \item We propose a novel HGS framework for compact and efficient dynamic scene reconstruction, explicitly separating static and dynamic components within a unified rendering and optimization pipeline.

    \item We introduce the SDD strategy, which employs parameter sharing by assigning a compact set of temporally invariant parameters to all static primitives. Combined with explicit modeling, SDD significantly reduces model complexity, thereby achieving faster rendering while maintaining compatibility with dynamic modeling.

    \item We design a two-stage training strategy specifically for our explicit Gaussian representation. This iterative approach enhances temporal consistency and effectively reduces artifacts at static-dynamic boundaries through mutual adaptation.

    \item As shown in Fig.~\ref{fig:teaser}, our approach delivers rendering quality comparable to state-of-the-art methods while reducing model size by up to 98\%, yielding a far more compact and efficient representation. It also better preserves high-frequency details in static regions and suppresses temporal artifacts under abrupt scene changes.
\end{itemize}

\section{Related Works}
\subsection{Dynamic NeRF}
Recent progress in dynamic NeRF~\cite{mildenhall2021nerf} has primarily focused on extending static scene methods to handle temporal dynamics. Many studies leverage deformation-based methods~\cite{guo2023forward, li2021neural, park2021nerfies, park2021hypernerf, pumarola2021d, tretschk2021non, xian2021space}, introducing neural deformation fields to warp dynamic scenes into a canonical static frame. For example, methods such as D-NeRF~\cite{pumarola2021d}, Nerfies~\cite{park2021nerfies}, and HyperNeRF~\cite{park2021hypernerf} effectively capture complex scene dynamics through learned deformations but incur substantial computational costs due to dense volumetric sampling. To alleviate these computational issues, structured representation methods~\cite{cao2023hexplane, fang2022fast, fridovich2023k, shao2023tensor4d, wang2023mixed, wang2023neural, song2023nerfplayer, attal2023hyperreel} have emerged. K-Planes~\cite{fridovich2023k} and MixVoxels~\cite{wang2023mixed} factorize the spacetime domain into structured planes or adaptive voxel grids, significantly reducing computational complexity. NeRFPlayer~\cite{song2023nerfplayer} integrates explicit time decomposition to better handle dynamic variations, while HyperReel~\cite{attal2023hyperreel} proposes low-dimensional temporal embeddings to enhance efficiency. However, these structured methods may still struggle to represent rapid or highly intricate motion accurately.
\subsection{Dynamic 3DGS}
The substantial advancements in rendering quality and efficiency brought by 3DGS have motivated subsequent research~\cite{huang2024sc, jeong2024rodygs, kratimenos2024dynmf, liang2023gaufre, wu20244d, yang2024deformable, lin2024gaussian, li2024spacetime, duan20244d, yang2023real, luiten2024dynamic, bae2024per}, aimed at extending the originally static 3DGS framework to dynamic scene reconstruction. Early works~\cite{luiten2024dynamic, wu20244d} employ MLPs to regress per-frame deformations of Gaussian parameters, enabling flexibility but incurring significant computational overhead. ED3DGS~\cite{bae2024per} enhances this formulation by introducing per-Gaussian latent embeddings for deformation and decomposing motion into coarse and fine levels. RoDyGS~\cite{jeong2024rodygs} incorporates a motion basis and separates static and dynamic points during the training of deformation fields to improve robustness and efficiency. However, all of these methods rely on implicit motion modeling, which limits rendering speed and increases model complexity.

To overcome the limitations of implicit modeling, 4DGS~\cite{yang2023real} extends 3D Gaussians into 4D space by attaching temporal features to each primitive, while STGS~\cite{li2024spacetime} models the temporal evolution of positions and orientations via polynomial functions and the temporal RBF. These methods improve runtime efficiency and avoid the need for implicit deformation fields. However, they often treat all Gaussian primitives as dynamic, including those belonging to static regions, leading to parameter redundancy and potential temporal blurring.

Unlike prior works that rely on implicit deformation fields to model dynamic motion, our approach adopts a temporal RBF that avoids such computational overhead. Additionally, we introduce a SSD strategy, assigning time-dependent parameters only to dynamic regions while modeling static regions with a shared, time-independent representation. This design reduces parameter redundancy, avoids temporal artifacts in static regions, and enables significantly faster rendering while maintaining visual quality comparable to state-of-the-art methods.

\begin{figure*}[t]
 \centering 
 \includegraphics[width=\textwidth]{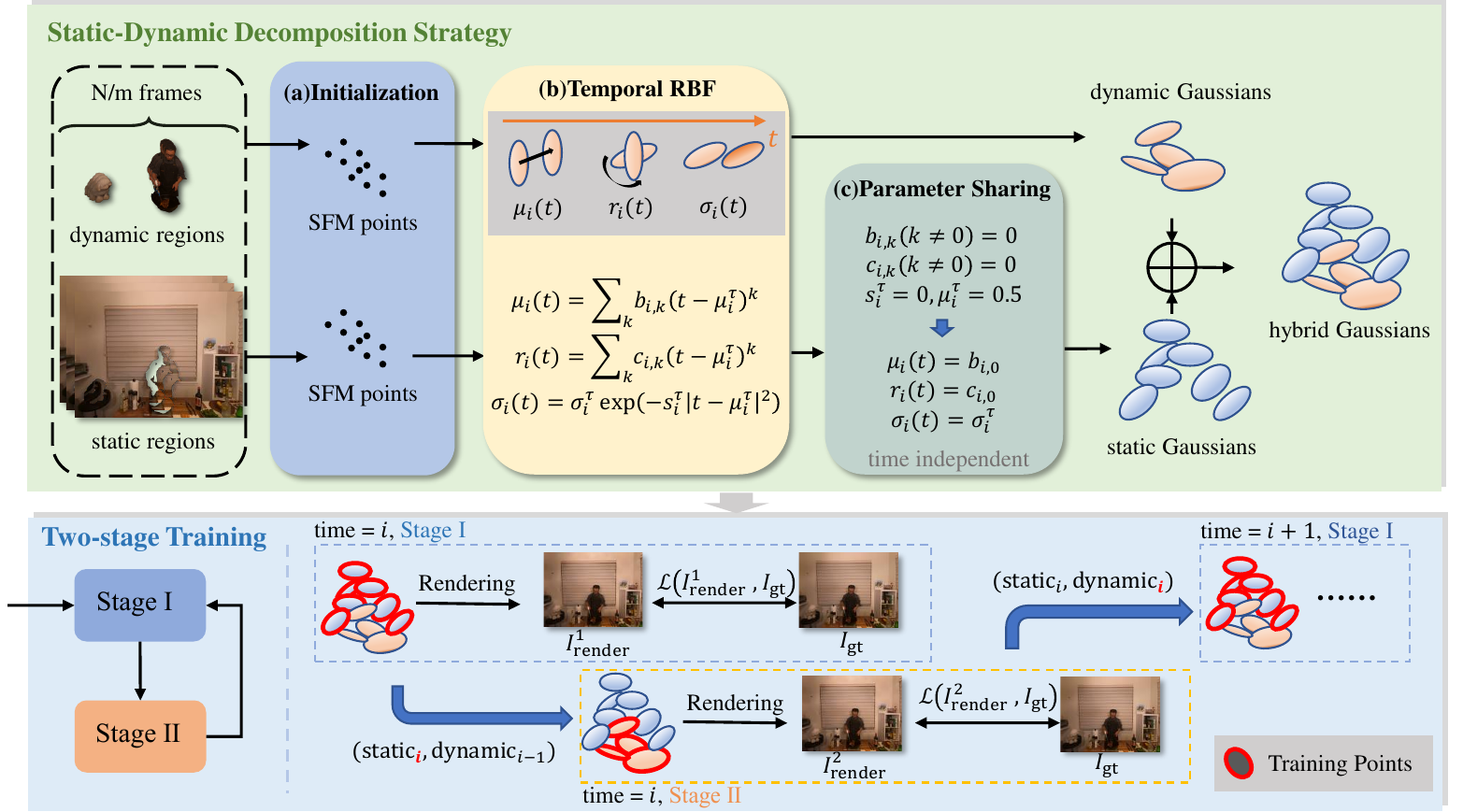}
 \caption{Overview of the proposed HGS pipeline, which comprises two components: (top) Static-Dynamic Decomposition Strategy, including (a) Initialization, selecting every $m$ th frame for initialization to reduce the number of SFM points; (b) Temporal RBF, adding time-dependent parameters for dynamic Gaussian primitives; and (c) Parameter Sharing, enforcing time-independent parameters for static Gaussian primitives. (bottom) Two-Stage Training: Stage I updates static Gaussian primitives, while Stage II jointly optimizes dynamic Gaussian primitives using the refined static primitives from Stage I.}
 \label{fig:pipeline}
\end{figure*}

\section{Preliminary: 3D Gaussian Splatting}
3DGS~\cite{kerbl20233d} is an explicit representation method that models scenes with differentiable anisotropic 3D Gaussians. Given a set of cali
brated images captured from multiple views, 3DGS optimizes Gaussian parameters of these Gaussians via differentiable rasterization to rep
resent scenes effectively and render novel views efficiently.

Specifically, each Gaussian $G_i$ in a scene is parameterized by several attributes: 
a spatial position $\mu_i \in \mathbb{R}^3$, 
a covariance matrix $\Sigma_i \in \mathbb{R}^{3 \times 3}$ that defines its shape and orientation, 
an opacity scalar $\sigma_i \in \mathbb{R}$ controlling transparency, 
and spherical harmonics (SH) coefficients $\mathbf{h}_i$ encoding view-dependent appearance.


The 3D Gaussian kernel at any spatial location $x\in \mathbb{R}^3$ can be formally expressed as:
\begin{equation}
\alpha_i(x) = \sigma_i \exp\left(-\frac{1}{2}(x-\mu_i)^{T}\Sigma_i^{-1}(x-\mu_i)\right),
\label{eq:3dgs}
\end{equation}
where the covariance matrix $\Sigma_i$ is symmetric and positive semi-definite. 
To ensure these constraints and achieve intuitive geometric interpretation, 
$\Sigma_i$ is decomposed using scaling and rotation components:
\begin{equation}
\Sigma_i = R_i S_i S_i^{T} R_i^{T},
\label{eq:cov}
\end{equation}
with $S_i$ being a diagonal scaling matrix parameterized by a scaling vector $s_i \in \mathbb{R}^3$ 
and $R_i \in \mathbb{R}^{3 \times 3}$ a rotation matrix derived from a quaternion representation 
$q_i \in \mathbb{R}^4$, enforcing unit-norm constraint for stability.

To render an image from a specific viewpoint, each 3D Gaussian is projected from the 3D world coordinate system 
onto the 2D image plane via an approximate perspective projection. 
Given camera extrinsics (viewing matrix) $W \in \mathbb{R}^{4 \times 4}$ 
and intrinsics (projection matrix) $K \in \mathbb{R}^{3 \times 4}$, 
the 3D position $\mu_i$ and covariance $\Sigma_i$ are projected to obtain the corresponding 
2D Gaussian distribution with mean $\mu^{2D}_i$ and covariance matrix $\Sigma^{2D}_i$:
\begin{align}
\mu_{i}^{2D} &= \left(K\frac{W\mu_i}{(W\mu_i)_z}\right)_{1:2},\\[5pt]
\Sigma_{i}^{2D} &= (J W \Sigma_i W^T J^T)_{1:2,1:2},
\end{align}
where $(W\mu_i)_z$ denotes the depth after viewing transformation, 
and $J$ is the Jacobian matrix of the affine approximation of the perspective projection, reflecting the local linearization of the nonlinear projective transformation.

After obtaining the projected Gaussians, rendering is achieved through alpha compositing along the camera viewing direction. The Gaussians are first sorted in descending order based on their depth relative to the camera. 
The final pixel color $I$ at the pixel coordinate is computed through alpha-blending of sorted Gaussians that influence the pixel:
\begin{equation}
I = \sum_{i \in N} c_i \alpha_i^{2D} \prod_{j=1}^{i-1}(1-\alpha_j^{2D}),
\end{equation}
where $N$ denotes the ordered set of Gaussians influencing the pixel, and the opacity $\alpha^{2D}_i$ at the pixel position $x_{2D}$ is evaluated as the 2D counterpart of the 3D Gaussian kernel:
\begin{equation}
\alpha_i^{2D}(x_{2D}) = \sigma_i \exp\left(-\frac{1}{2}(x_{2D}-\mu_i^{2D})^{T}(\Sigma_i^{2D})^{-1}(x_{2D}-\mu_i^{2D})\right),
\end{equation}
and the color $c_i$ is computed from spherical harmonics(SH) coefficients $h_i$ given the corresponding viewing direction $d_v$:
\begin{equation}
c_i = \text{SH}(h_i, d_v),
\label{eq:color}
\end{equation}
where $\text{SH}(\cdot)$ denotes the evaluation of spherical harmonics.

\section{Method}
In this section, we introduce our HGS framework for dynamic scene representation with lower storage cost, higher rendering speed, and high quality. Our method achieves this by incorporating an SDD strategy and a two-stage training strategy. This design not only enables a more compact representation but also supports efficient, high-quality rendering. An overview of our method is illustrated in Fig.~\ref{fig:pipeline}. Sec.~\ref{sec:HGS} describes the hybrid Gaussian primitives. Sec.~\ref{sec:SDD} presents our SDD strategy for detailed modeling. Sec.~\ref{sec:training} introduces the two-stage training strategy, followed by optimization details in Sec.~\ref{sec:opti}.

\subsection{Hybrid Gaussians}
\label{sec:HGS}
To efficiently and compactly represent dynamic 4D scenes, we explicitly separate static and dynamic primitives within a unified rendering and optimization framework. Inspired by recent work on STGS~\cite{li2024spacetime}, our approach extends traditional 3D Gaussian primitives into the 4D spacetime domain by explicitly modeling their temporal evolution. Specifically, each Gaussian primitive $G_i$ is parameterized by temporally varying spatial position $\mu_i(t)$, opacity $\sigma_i(t)$, rotation quaternion $r_i(t)$, as well as temporally invariant anisotropic scale vector $s_i$ and color coefficients $c_i$. The explicit formulations for these parameters are given in Eq. (\ref{eq:3dgs}), Eq. (\ref{eq:cov}), and Eq. (\ref{eq:color}), respectively. Within this modeling framework, we propose a novel approach that separates static and dynamic primitives using a binary indicator, and applies distinct parameterization strategies for each category of primitives.

\subsubsection{Dynamic Primitives} For dynamic primitives, each Gaussian independently maintains its own temporally varying parameters. The temporal evolution of each dynamic primitive is explicitly parameterized. Specifically, each Gaussian parameter is modeled through polynomial interpolation. The detailed formulations are described in Sec.~\ref{sec:rbfs}.

\subsubsection{Static Primitives} All static Gaussian primitives share a unified, temporally invariant parameter set. By leveraging this shared parameterization, our method significantly reduces redundancy, thereby providing a compact and efficient representation of static scene components. The detailed parameter-sharing strategy is described in Sec.~\ref{sec:share}.

\subsection{Static-Dynamic Decomposition Strategy}
\label{sec:SDD}
To realize this idea more effectively, we introduce the SDD strategy, which provides a concrete implementation of separate modeling for dynamic and static regions within our hybrid Gaussian framework. We first perform initialization to obtain Gaussian primitives for both categories. For dynamic regions, we apply temporal RBF~\cite{li2024spacetime} to capture time-varying behavior. In contrast, static regions share a single set of temporally invariant parameters, allowing for efficient modeling without the overhead of redundant dynamic Gaussians.

\subsubsection{Initialization} To enable compact and efficient scene representation, we adapt the interleaved-frame initialization scheme. Rather than relying on dense SfM point clouds from all frames, hybrid Gaussian primitives are initialized using reconstructions from temporally subsampled frames. This approach limits the number of initial Gaussian primitives, ensuring a favorable trade-off between computational efficiency and the fidelity needed to capture the scene.

\subsubsection{Radial Basis Function}\label{sec:rbfs}
For dynamic primitives, we adapt the temporal RBF of STGS~\cite{li2024spacetime} to explicitly parameterize the temporal evolution. The formulations are as follows:
\begin{align}
        \mu_i(t) &= \sum_{k=0}^{3} b_{i,k}(t - \mu_i^\tau)^k,\\[5pt]
        r_i(t) &= \sum_{k=0}^{1} c_{i,k}(t - \mu_i^\tau)^k,\\[5pt]
        \sigma_i(t) &= \sigma_i^{\tau}\exp\left(-s_i^\tau|t - \mu_i^\tau|^2\right),
\end{align}
where $\mu_i(t)$ and $r_i(t)$ explicitly encode trajectories of position and rotation through polynomial interpolation. Here, $b_{i,k}$ and $c_{i,k}$ denote the polynomial coefficients that are optimized during training, while the temporal center $\mu_i^\tau$ defines the central time around which each Gaussian’s trajectory is modeled. The opacity $\sigma_i(t)$ is formulated as a Gaussian-shaped temporal function, parameterized by a spatial opacity factor $\sigma_i^\tau$, a temporal scaling factor $s_i^\tau$, and the temporal center $\mu_i^\tau$. For simplicity and computational efficiency, both the scale $s_i$ and color $c_i$ are treated as time-invariant, inspired by~\cite{lee2024compact}.

Instead of relying on implicit motion modeling, such as deformation fields ~\cite{bae2024per, jeong2024rodygs}, which comes with a heavy parameter burden, our explicit modeling method uses fewer parameters while ensuring comparable rendering results. 

\subsubsection{Parameter Sharing}\label{sec:share}
To reduce parameter redundancy and improve model compactness, we implement a dedicated parameter-sharing strategy. Specifically, during initialization, static Gaussian primitives are generated directly from the SfM point clouds. For each static primitive $G_i$ in the scene, we explicitly set the temporal polynomial coefficients $b_{i,k}$ ($k=1,2,3$), the first-order rotation coefficient $c_{i,1}$, and the temporal scaling factor $s_i^\tau$ to zero, and set the temporal center $\mu_i^\tau$ to one-half. Thus, while static primitives formally share the same parametric structure as dynamic counterparts, they effectively degenerate into temporally invariant primitives.

Since these parameters are identical across all static primitives, it is unnecessary to store them individually during model serialization. Consequently, our parameter-sharing strategy substantially reduces parameter redundancy, significantly decreasing the model size while preserving representational accuracy and rendering efficiency.

By combining temporally fixed parameter sharing for static primitives with flexible temporal modeling for dynamic primitives, our SDD strategy achieves significant parameter reduction while retaining the flexibility to represent complex scene dynamics. The fixed parameter sharing for static primitives also eliminates cross-region temporal interference, preserving high-frequency details and visual fidelity within static content.

\subsection{Two-stage Training}
\label{sec:training}
Although the SDD strategy provides a compact and efficient representation, the explicit separation of static and dynamic regions can introduce artifacts near their boundaries. To address this issue, we propose a novel two-stage training strategy, specifically designed to facilitate smoother transitions and enhance consistency across static-dynamic boundaries. Our approach encourages dynamic primitives to better fit and adapt to static regions, significantly alleviating boundary artifacts and ensuring overall scene coherence.

During training, we decompose the Hybrid Gaussians into static and dynamic ones. When Gaussian primitives are split or generated during optimization, they inherit the static/dynamic labels of their parent primitives, ensuring coherence throughout the training process.

Within each optimization cycle, our two-stage training proceeds as follows:

\subsubsection{Stage 1: Static Optimization} We first optimize only the parameters of the static primitives. Specifically, we render the hybrid Gaussians derived from the previous optimization iteration and compute the rendering loss between the rendered images and the ground truth images. This loss is utilized to update the parameters of static Gaussian primitives exclusively, stabilizing the static representation and refining scene geometry and appearance.
    
\subsubsection{Stage 2: Dynamic Optimization} Subsequently, we merge the newly updated static primitives from Stage 1 with the previously optimized dynamic primitives, forming updated hybrid Gaussians. Rendering these hybrid Gaussians again, we compute the loss relative to the ground-truth images and use this loss to update only the parameters of the dynamic Gaussian primitives. Crucially, in this stage, the dynamic primitives adapt to the optimized static representation, further improving consistency at static-dynamic boundaries.

Compared to training static and dynamic primitives separately in an isolated manner, our sequential two-stage training approach effectively mitigates the boundary discontinuities and temporal blurring artifacts at the interfaces between static and dynamic regions. Although this iterative optimization slightly increases the total training time, it significantly enhances the seamless integration between static and dynamic primitives, leading to a coherent, high-quality reconstruction of dynamic scenes.

\begin{figure*}[t]
  \centering
  \includegraphics[width=0.9\textwidth]{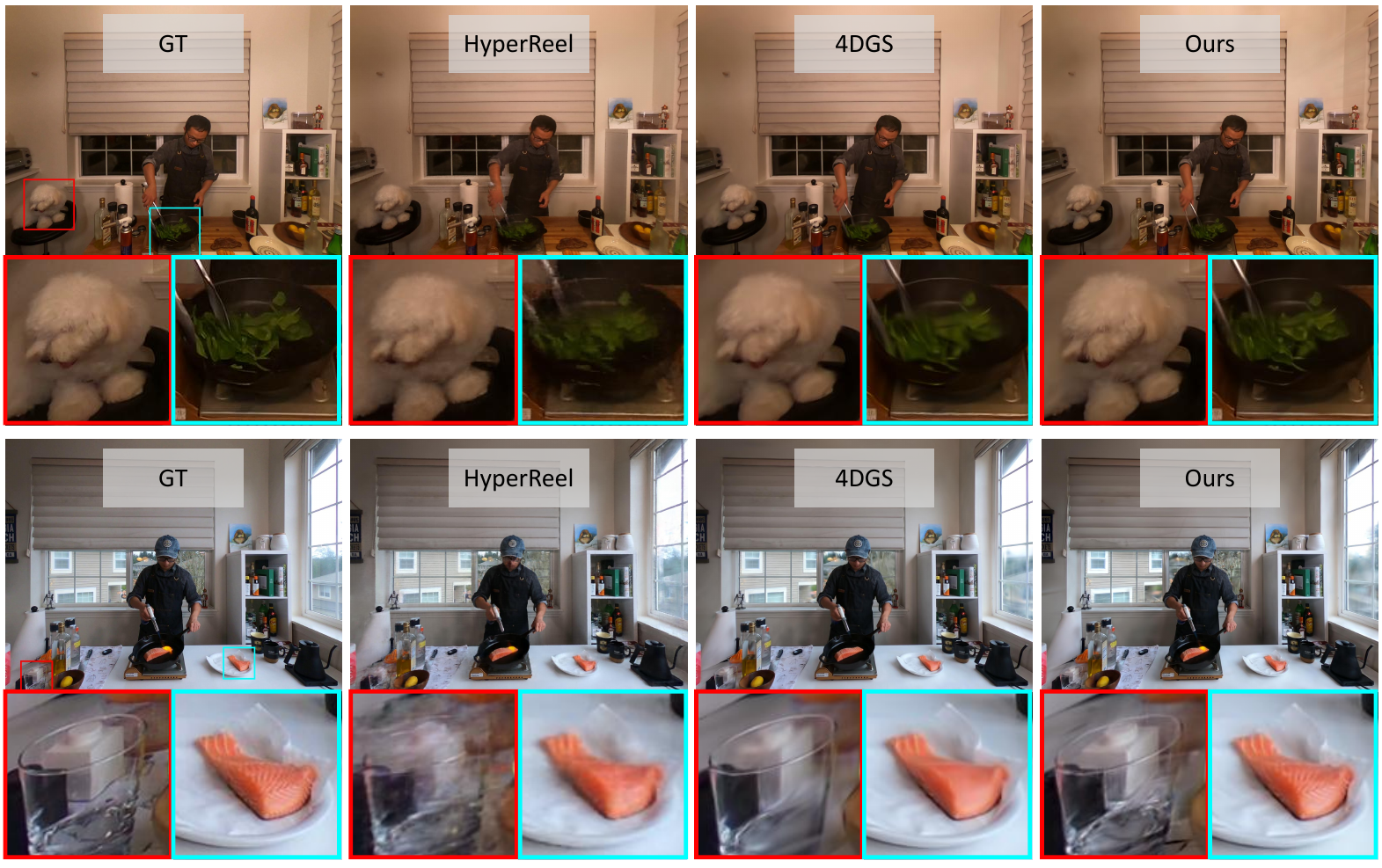}
  \includegraphics[width=0.9\textwidth]{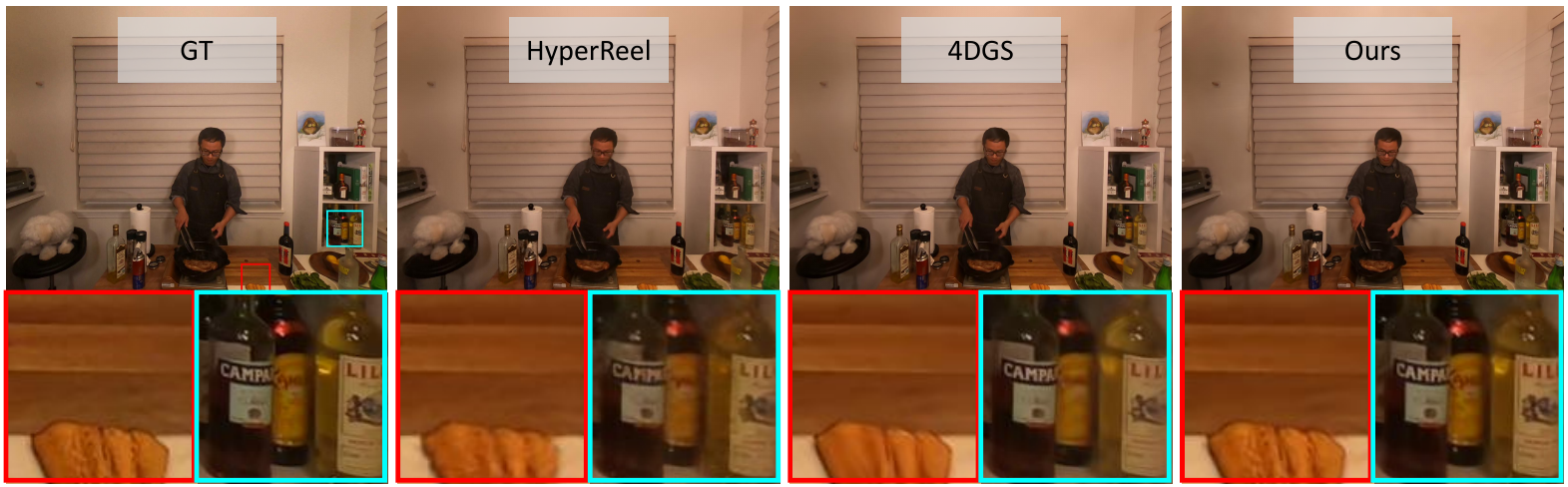}
  \caption{Qualitative results comparison on Neural 3D Video Dataset. The red and blue highlight areas where the proposed method achieves notable visual quality improvements.}
  \label{fig:n3d}
\end{figure*}

\subsection{Optimization}
\label{sec:opti}

With the modeling framework and two-stage training strategy in place, we now present the optimization of HGS parameters, performed via differentiable rasterization and gradient-based optimization. Due to the distinct nature of static and dynamic primitives in our framework, their parameter sets differ accordingly. Specifically, the optimization parameters for each static primitive include $\sigma_i^\tau$, $b_{i,0}$, $c_{i,0}$, $s_i$, and $c_i$. For each dynamic primitive, we optimize parameters controlling temporal variation, namely $\sigma_i^\tau$, $s_i^\tau$, $\mu_i^\tau$, polynomial trajectory coefficients $\{b_{i,k}\}_{k=0}^{3}$, rotation coefficients $\{c_{i,k}\}_{k=0}^{1}$, as well as anisotropic scale vector $s_i$ and color coefficients $c_i$.

The optimization process interleaves gradient-based parameter updates with Gaussian density control strategies such as splitting and pruning, ensuring that our model remains both compact and expressive throughout training.

Following previous methods~\cite{kerbl20233d}, we define the rendering loss as a combination of an $L_1$ photometric term and a structural similarity index (D-SSIM) term, which robustly captures visual differences between rendered and ground-truth images. Given a rendered image $I_{\mathrm{render}}$ and its corresponding ground-truth image $I_{\mathrm{gt}}$, the rendering loss $\mathcal{L}$ is formulated as follows:
\begin{equation}
    \mathcal{L}(I_{\mathrm{render}}, I_{\mathrm{gt}}) = \| I_{\mathrm{render}} - I_{\mathrm{gt}} \|_1 + \lambda \cdot \mathcal{L}_{\mathrm{D\text{-}SSIM}}(I_{\mathrm{render}}, I_{\mathrm{gt}}),
\end{equation}
where $\lambda$ balances the contribution between the two terms, and $\mathcal{L}_{\mathrm{D\text{-}SSIM}}$ denotes the D-SSIM loss term.

During training, we alternately optimize static and dynamic primitives following the proposed two-stage optimization strategy (detailed in Sec.~\ref{sec:training}), enabling effective modeling of dynamic interactions between scene components and improved consistency at static-dynamic boundaries.

\section{Experiments}
\subsection{Experimental Setup}
\subsubsection{Implementation Details} 
We employ the Track Anything Model (TAM)~\cite{yang2023track} to perform explicit separation between static and dynamic regions in the input videos. The resulting segmentation masks from TAM are saved and utilized to prepare our datasets for training. According to~\cite{navaneet2024compgs}, we employ 3DGS without the harmonic components for color to further reduce the number of parameters and achieve high-quality rendering. For optimization, we use the Adam optimizer~\cite{kingma2014adam}. All experiments and training are conducted on a single NVIDIA RTX 4090 GPU (24GB), except real-time rendering, which is performed on an RTX 3090 GPU (24GB).

\begin{figure*}[t]
  \centering
  \includegraphics[width=\textwidth]{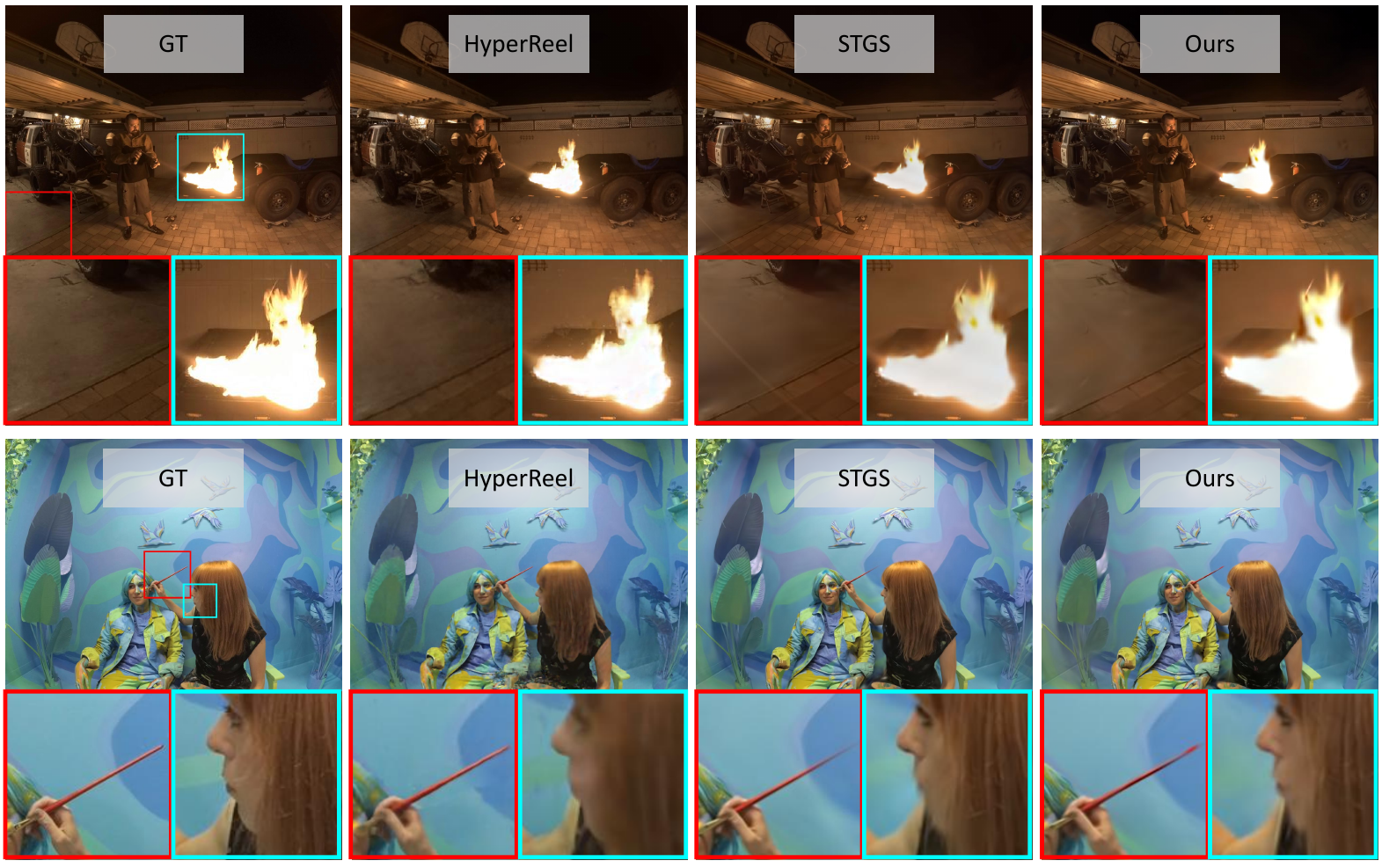}
  \caption{Qualitative results comparison on Google Immersive Dataset. The red and blue highlight areas where the proposed method achieves notable visual quality improvements.}
  \label{fig:immer}
\end{figure*}

\subsubsection{Datasets.} 
We evaluate our approach quantitatively and qualitatively on two public datasets: the Neural 3D Video Dataset~\cite{li2022neural} and the Google Immersive Dataset~\cite{broxton2020immersive}. The Neural 3D Video Dataset, captured using 19–21 cameras at 2704×2028 resolution and 30 FPS, contains dynamic scenes with complex motions and variable object appearances, posing challenges for consistent reconstruction. The Google Immersive Dataset, recorded with a 44–46 camera fisheye rig, includes indoor and outdoor scenes with abrupt transitions—such as the sudden appearance of a large fire source—further complicating multi-view reconstruction. Owing to the inherent challenges of the Google Immersive Dataset, only a few works have attempted comparative evaluations on it.

\subsubsection{Comparison Methods.} 
We compare our method with recent dynamic NeRF-based and Gaussian-based approaches for dynamic scene reconstruction from multi-view videos. The NeRF-based baselines include K-Planes~\cite{fridovich2023k}, MixVoxels~\cite{wang2023mixed},  HyperReel~\cite{attal2023hyperreel}, and NerfPlayer~\cite{song2023nerfplayer}, while the Gaussian-based baselines include 4DGS~\cite{wu20244d}, ED3DGS~\cite{bae2024per}, STGS~\cite{li2024spacetime} and SaRO-GS~\cite{yan20244d}. For all comparisons, we use the official implementations and keep the original hyperparameters unchanged.

\subsection{Results}
\subsubsection{Quantitative Comparison}
As shown in Tab.~\ref{tab:comparison}, our method achieves a PSNR of 32.36 dB and an SSIM of 0.952, outperforming strong baselines~\cite{attal2023hyperreel, fridovich2023k, wang2023mixed, song2023nerfplayer, wu20244d, bae2024per} while attaining visual quality comparable to STGS~\cite{li2024spacetime} and offering significantly improved efficiency. Although the rendering quality is marginally lower than that of SaRO-GS~\cite{yan20244d}, our approach offers notable benefits in terms of model compactness, training efficiency, and rendering speed. In particular, our model requires only 6.87 MB of storage, marking a 63\% reduction compared to STGS~\cite{li2024spacetime} and up to a 98\% reduction relative to MixVoxels~\cite{wang2023mixed}. These results attest to the effectiveness of our design in achieving a compact yet efficient dynamic scene representation without sacrificing visual quality. Furthermore, our method consistently performs well across all test scenes, as detailed in the supplementary material.
\begin{table}[t]
    \centering
    \caption{Quantitative comparisons on the Neural 3D Video Dataset.}
    \label{tab:comparison}
    \resizebox{\linewidth}{!}{
    \begin{tabular}{lcccccc}
        \toprule
        \textbf{Method} & \textbf{PSNR}$\uparrow$ & \textbf{SSIM}$\uparrow$ & \textbf{Storage(MB)}$\downarrow$ & \textbf{Training time}$\downarrow$ & \textbf{FPS}$\uparrow$ \\
        \midrule
        K-Planes~\cite{fridovich2023k} & 31.39 & 0.944 & 311 &  -    & 0.3      \\
        MixVoxels~\cite{wang2023mixed}         & 30.55 & 0.924 & 508.07 & 1h 22m     &  16.7   \\
        HyperReel~\cite{attal2023hyperreel}     & 31.65 & 0.934 & 169.47 & 1h 41m & 2.00   \\
        4DGS~\cite{wu20244d}     & 32.02 & 0.946 & 29.36 & 31m & 114    \\
        ED3DGS~\cite{bae2024per}     & 32.35 & 0.949 & 31.41 & 1h 35m & 75    \\
        STGS~\cite{li2024spacetime}     & 32.32 & \textcolor{blue}{0.953} & \textcolor{blue}{18.19} & \textcolor{blue}{29m} & \textcolor{blue}{140}    \\
        SaRO-GS~\cite{yan20244d}     & \textcolor{red}{33.06} & \textcolor{red}{0.955} & 236.44 & 1h 31m & 40    \\
        Ours                     & \textcolor{blue}{32.36} & 0.952 & \textcolor{red}{6.87} & \textcolor{red}{18m} & \textcolor{red}{300+}  \\
        \bottomrule
    \end{tabular}
    }
\end{table}
\begin{table}[t]
    \centering
    \caption{Quantitative comparisons on the Google Immersive Dataset.}
    \label{tab:immersive}
    \resizebox{\linewidth}{!}{
    \begin{tabular}{lcccccc}
        \toprule
        \textbf{Method} & \textbf{PSNR}$\uparrow$ & \textbf{SSIM}$\uparrow$ & \textbf{Storage(MB)}$\downarrow$ & \textbf{FPS}$\uparrow$ \\
        \midrule
        NeRFPlayer~\cite{song2023nerfplayer}         & 27.23 & 0.887 & 5130 &  0.12    \\
        HyperReel~\cite{attal2023hyperreel}         & \textcolor{blue}{29.09} & 0.899 & 190.68 &  4    \\
        STGS~\cite{li2024spacetime}        & 28.87 & \textcolor{red}{0.943} & \textcolor{blue}{19.74} &  \textcolor{blue}{99}    \\
        Ours      & \textcolor{red}{29.60} & \textcolor{blue}{0.925} & \textcolor{red}{12.73} &  \textcolor{red}{300+}    \\
        \bottomrule
    \end{tabular}
    }
\end{table}

In terms of computational efficiency, our extensive experiments show that HGS delivers exceptional real-time performance, achieving speed over 300 FPS at a resolution of $1352 \times 1014$ on a single NVIDIA RTX 3090 GPU, far surpassing STGS~\cite{li2024spacetime} (140 FPS) and ED3DGS~\cite{bae2024per} (75 FPS). Moreover, HGS significantly improves training efficiency by completing training in only 18 minutes, considerably faster than competing methods such as SaRO-GS~\cite{yan20244d}, which requires 1 hour and 31 minutes. This efficient pipeline further highlights the practicality and scalability of our approach for real-world dynamic scene applications.

Tab.~\ref{tab:immersive} summarizes quantitative results on the Google Immersive Dataset. Characterized by abrupt scene transitions, our method attains a competitive PSNR of 29.60 dB and an SSIM of 0.925, outperforming baselines like HyperReel~\cite{attal2023hyperreel} and STGS~\cite{li2024spacetime}. To mitigate artifacts arising from sudden transitions, our approach supplements Gaussians, resulting in a modest increase in model size. Detailed visual results are provided in the qualitative comparison for further insight.

\subsubsection{Qualitative Comparison.}
To qualitatively evaluate the visual fidelity of our proposed method, we conduct comparisons on the two aforementioned datasets. On the Neural 3D Video Dataset, we visualize reconstruction results from two representative scenes, emphasizing fidelity in high-frequency texture regions. As shown in Fig.~\ref{fig:n3d}, compared to the two baseline methods~\cite{attal2023hyperreel, wu20244d}, our method produces reconstructions significantly closer to the ground truth, particularly preserving intricate textures such as the detailed surface of meat and the complex layering of vegetable leaves. This confirms our method's effectiveness in addressing the commonly observed issue of missing high-frequency details in static regions.

On the Google Immersive Dataset, we focus on artifacts induced by abrupt scene changes. As shown in Fig.~\ref{fig:immer}, when a sudden flame brightens the scene, STGS~\cite{li2024spacetime} fails to capture the brightness accurately, resulting in artifacts in static regions. Similar issues occur in the reconstruction of moving objects for both STGS~\cite{li2024spacetime} and HyperReel~\cite{attal2023hyperreel}. In contrast, our method mitigates these artifacts, handling sudden illumination shifts and abrupt content changes.

Overall, our qualitative analysis shows that our approach preserves high-frequency texture details and mitigates artifacts from abrupt scene transitions, resulting in superior visual quality. For more qualitative comparisons with state-of-the-art methods, please refer to the appendix.

\subsection{Ablation Study}
\begin{table}[t]
\centering
\caption{Quantitative results of the ablation study on our framework, reported as averages over the Neural 3D Video Dataset.}
\label{tab:ablation}
\resizebox{\linewidth}{!}{
\begin{tabular}{lccc}
\toprule
\textbf{Method} & \textbf{PSNR}↑ & \textbf{SSIM}↑ & \textbf{Storage(MB)}↓ \\
\midrule
w/o Initialization      & 32.5 & 0.954 & 13.51 \\
w/o Parameter Sharing        & 32.22 & 0.951 & 11.56 \\
w/o Two-stage Training         & 25.76 & 0.910 & 5.88 \\
Ours           & 32.36 & 0.952 & 6.87 \\
\bottomrule
\end{tabular}
}
\end{table}

\begin{figure*}[t]
  \centering
  \includegraphics[width=0.9\textwidth]{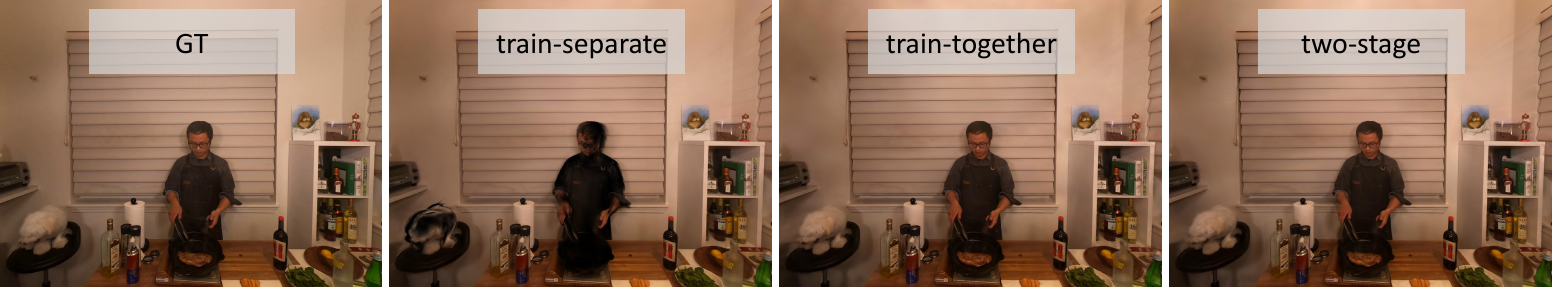}
  \caption{Qualitative results of the Training Strategies for Static and Dynamic
Regions.}
  \label{fig:three-train}
\end{figure*}

\begin{figure*}[t]
  \centering
  \includegraphics[width=\textwidth]{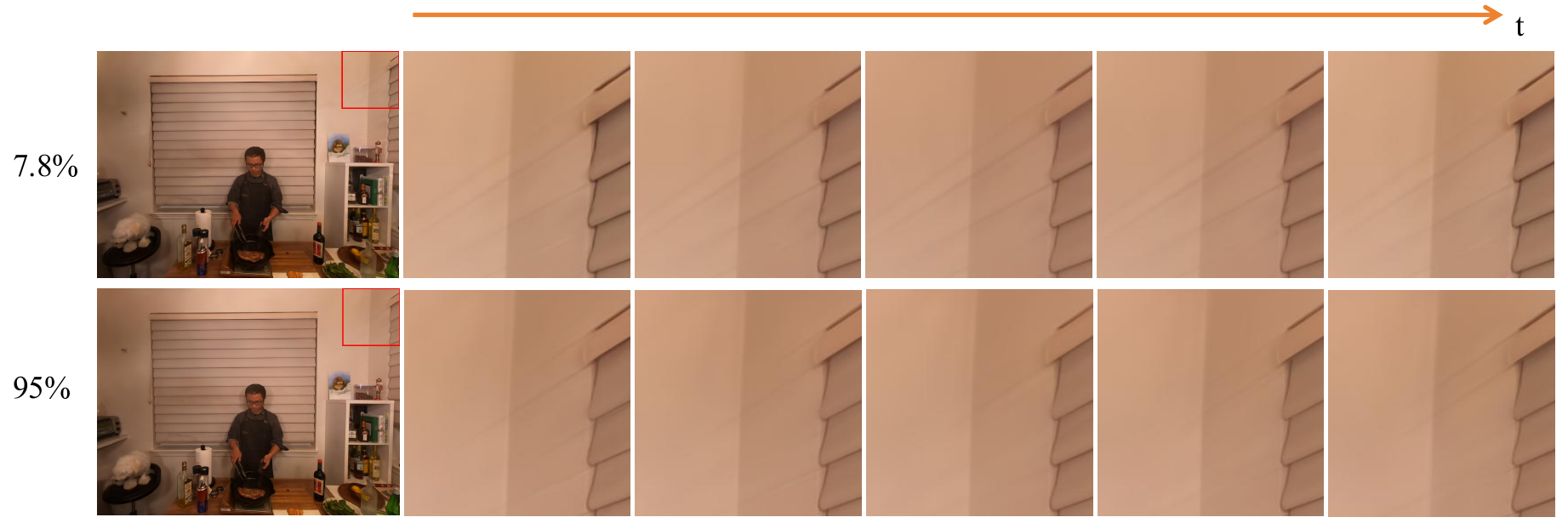}
  \caption{Qualitative results of the Analysis of Dynamic Region Proportion. The 7.8\% setting corresponds to the dynamic region obtained through standard static-dynamic decomposition, while the 95\% setting is generated by dilating the dynamic mask to artificially expand the dynamic area.}
  \label{fig:dyn-prop}
\end{figure*}

To investigate the effectiveness of individual components within our framework, we conduct comprehensive ablation studies summarized in Tab.~\ref{tab:ablation}. In the following, we describe the configuration and performance of each ablation baseline.

\textbf{Initialization}: Omitting our interleaved-frame initialization increases storage size further to 13.51 MB. While PSNR slightly improves, this minor gain is outweighed by a substantial increase in model complexity. This highlights that our initialization strategy provides an optimal balance between rendering quality and model compactness.

\textbf{Parameter Sharing}: Removing the parameter-sharing strategy results in an increase in storage from 6.87 MB to 11.56 MB, nearly doubling the parameter count without any significant improvement in visual metrics. This underscores the effectiveness of our parameter-sharing strategy in dramatically reducing redundancy.

\begin{figure}[t]
  \centering
  \includegraphics[width=0.45\textwidth]{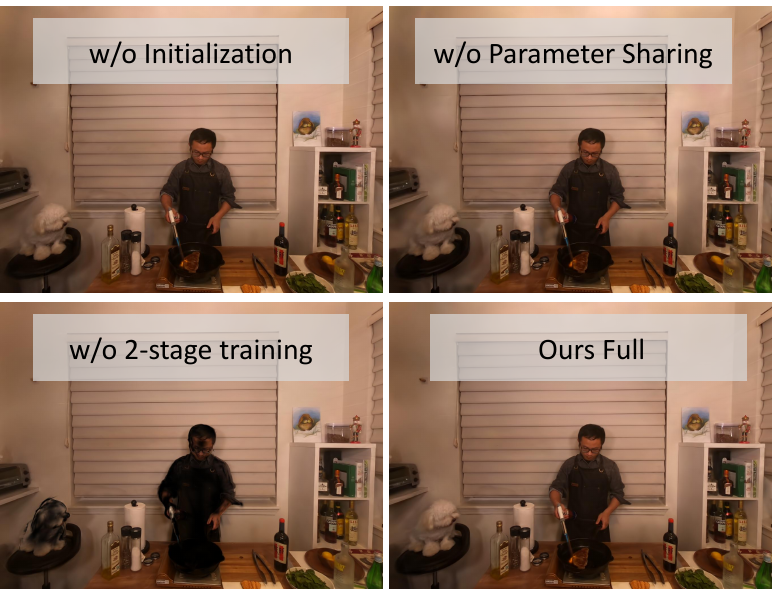}
  \caption{Qualitative results of the ablation study.}
  \label{fig:ablation}
\end{figure}
\textbf{Two-stage Training}: Removing our two-stage training strategy (i.e., training static and dynamic Gaussians separately without mutual adaptation) significantly reduces rendering quality, with PSNR dropping from 32.36 dB to 25.76 dB and SSIM from 0.952 to 0.910. This decline underscores the critical role of mutual adaptation between static and dynamic regions, facilitated by our two-stage strategy, in enhancing temporal consistency and reducing boundary artifacts. Although this iterative optimization slightly increases model size, it is justified by the considerable visual improvements demonstrated in Fig.~\ref{fig:ablation}.

\subsubsection{RBF Order}
To justify our choice of a third-order RBF, we conducted experiments with different polynomial orders. The results are summarized in the Tab.~\ref{tab:rbf-order}. The results demonstrate that selecting an RBF order of 3 provides the best overall balance. Compared to order 2, it achieves higher PSNR without sacrificing FPS, while only slightly increasing storage. Compared to order 5, it delivers nearly the same PSNR but with significantly lower storage requirements and faster FPS. Therefore, order 3 offers the most practical trade-off between reconstruction quality, efficiency, and resource consumption, justifying its use as the default choice in our method.
\begin{table}[t]
\centering
\caption{Results of different RBF orders evaluated on the Neural 3D Video Dataset.}
\label{tab:rbf-order}
\resizebox{\linewidth}{!}{
\begin{tabular}{lcccc}
\toprule
\textbf{Order} & \textbf{PSNR}↑ & \textbf{Storage(MB)}↓ & \textbf{Training time}↓ & \textbf{FPS}↑ \\
\midrule
2      & 32.16 & 6.74  & 18m & 300+ \\
3 (Ours)        & 32.36 & 6.87  & 18m & 300+\\
5        & 32.44 & 7.61  & 18m & 290\\
\bottomrule
\end{tabular}
}
\end{table}

\subsubsection{Variants of Training Strategies for Static and Dynamic Regions}
For training static and dynamic regions, there are three intuitive designs: training them separately, training them together (i.e., updating the parameters of static and dynamic primitives simultaneously), and our proposed alternating two-stage strategy. We conducted experiments under all three settings to validate the effectiveness of our design. The results are shown in Tab.~\ref{tab:two-stage-abl}. First, training static and dynamic regions separately leads to severe artifacts and a notable drop in quality. The results also clearly demonstrate that training static and dynamic blobs simultaneously is inefficient: the train-together strategy nearly doubles the storage cost while providing no gain in PSNR and even slightly underperforming the two-stage approach. In contrast, the two-stage strategy achieves higher PSNR with less storage and faster convergence while maintaining real-time FPS. This indicates that gradient interference between static and dynamic blobs hampers convergence when trained together, leading to inefficiency and potential overfitting. Therefore, adopting a two-stage training scheme is a more effective and practical solution. The qualitative results are shown in Fig.~\ref{fig:three-train}.

\begin{table}[t]
\centering
\caption{We tested the performance of three training methods on the Neural 3D Video Dataset.}
\label{tab:two-stage-abl}
\resizebox{\linewidth}{!}{
\begin{tabular}{lcccc}
\toprule
\textbf{Method} & \textbf{PSNR}↑ & \textbf{Storage(MB)}↓ & \textbf{Training time}↓ & \textbf{FPS}↑ \\
\midrule
train-separate     & 25.76 & 5.88  & 18m & 300+ \\
train-together     & 32.33 & 12.01  & 18m & 300+\\
two-stage (Ours)        & 32.36 & 6.87  & 18m & 300+\\
\bottomrule
\end{tabular}
}
\end{table}

\subsection{Effect of Dynamic Region Proportion on Performance}
\begin{table}[t]
\centering
\caption{Evaluation of the impact of increasing dynamic region proportion on the “Sear Steak” scene through progressive expansion of the dynamic mask.}
\label{tab:dyn-prop}
\resizebox{\linewidth}{!}{
\begin{tabular}{lccccc}
\toprule
\textbf{Proportion} & \textbf{PSNR}↑ & \textbf{LPIPS}↓ & \textbf{Storage(MB)}↓ & \textbf{Training time}↓ & \textbf{FPS}↑ \\
\midrule
7.8\% (Ours)      & 33.34 & 0.032 & 4.77 & 18m 31s & 149 \\
20\%        & 33.32 & 0.032 & 5.35 & 18m 30s & 136\\
40\%        & 33.44 & 0.032 &  6.18 & 19m 06s& 130\\
60\%        & 33.95 & 0.034 &  6.88 & 18m 11s & 131\\
80\%        & 33.61 & 0.033 &  7.73 & 19m 07s & 131\\
95\%        & 33.85 & 0.032 &  8.30 & 19m 21s & 132\\
\bottomrule
\end{tabular}
}
\end{table}

We assess the impact of dynamic region proportions on the performance of our method. To further highlight performance differences under the refresh-rate limitation of conventional displays, we increase the rendering resolution, which results in distinguishable variations in frame rates. Starting from the baseline where our method identifies only 7.8\% of the scene as dynamic, we gradually expand this proportion up to 95\% to stress-test performance. As shown in Tab.~\ref{tab:dyn-prop}, our method maintains stable PSNR and low LPIPS across varying dynamic proportions, reflecting reliable quantitative performance. However, when dynamic modeling is applied to regions that are inherently static, we observe that the slight numerical gains come at the cost of perceptual quality: artifacts such as dynamic blur and temporal flickering emerge, which noticeably reduce visual stability during immersive viewing. These artifacts not only increase storage demands but also distract users in real-time rendering, where consistent and artifact-free perception is far more critical than marginal improvements in objective scores. Representative examples are provided in the Fig.~\ref{fig:dyn-prop}. Additional demonstrations can be found in the supplementary video materials.

\begin{figure}[t]
  \centering
  \includegraphics[width=0.35\textwidth]{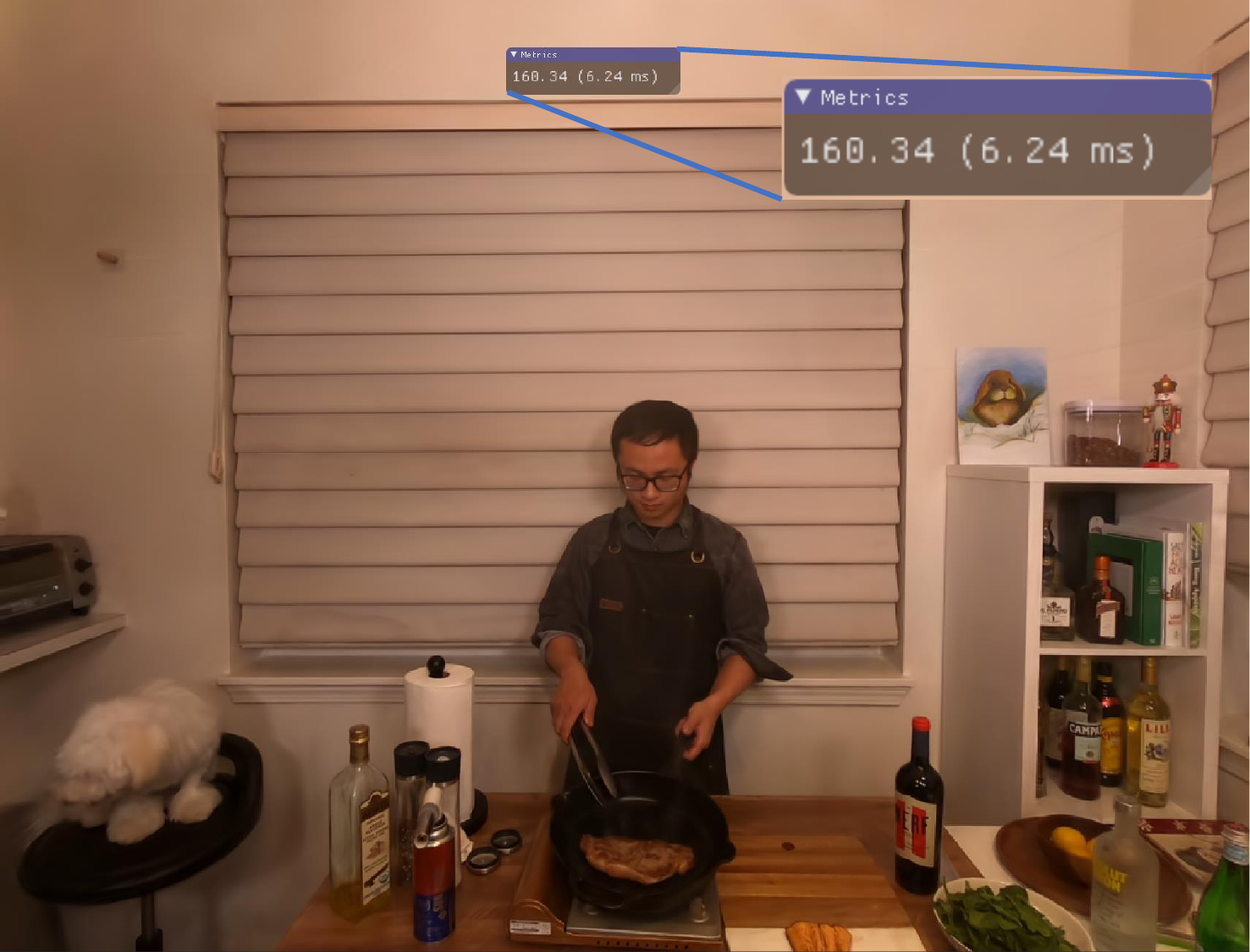}
  \caption{Real-time rendering performance on resource-constrained devices, validated on a laptop equipped with an RTX 3050 GPU.}
  \label{fig:low-comp}
\end{figure}

\subsection{Real-Time Rendering on Resource-Constrained Devices}
Prior works typically evaluate efficiency on high-end GPUs, neglecting the broader applicability of their methods in resource-constrained environments. In practice, however, the majority of users rely on mid- or low-tier devices, where achieving real-time rendering is far more challenging. To this end, we validate our framework on a laptop equipped with an RTX 3050 GPU, achieving real-time $1352 \times 1014$ rendering as shown in Fig.~\ref{fig:low-comp}. This result highlights the practical advantage of our method, which remains efficient even under strict hardware constraints. Regrettably, existing baselines do not provide real-time rendering players, making a direct runtime comparison impossible. Nevertheless, under the same resolution, our rendering frame rate on the RTX 3050 already surpasses that of all existing methods running on an RTX 3090, clearly demonstrating the superior performance of our approach on resource-limited devices.

\subsection{Real-Time Rendering in VR Systems}
Compared to conventional 2D rendering, VR rendering imposes far stricter computational demands, since it requires higher resolutions and stereo rendering for both eyes. These requirements place a heavy burden on the rendering pipeline. Existing dynamic-Gaussian methods, with their large model sizes and high complexity, struggle to meet such constraints and, to our knowledge, have not demonstrated VR applications.

\begin{figure}[t]
  \centering
  \includegraphics[width=0.45\textwidth]{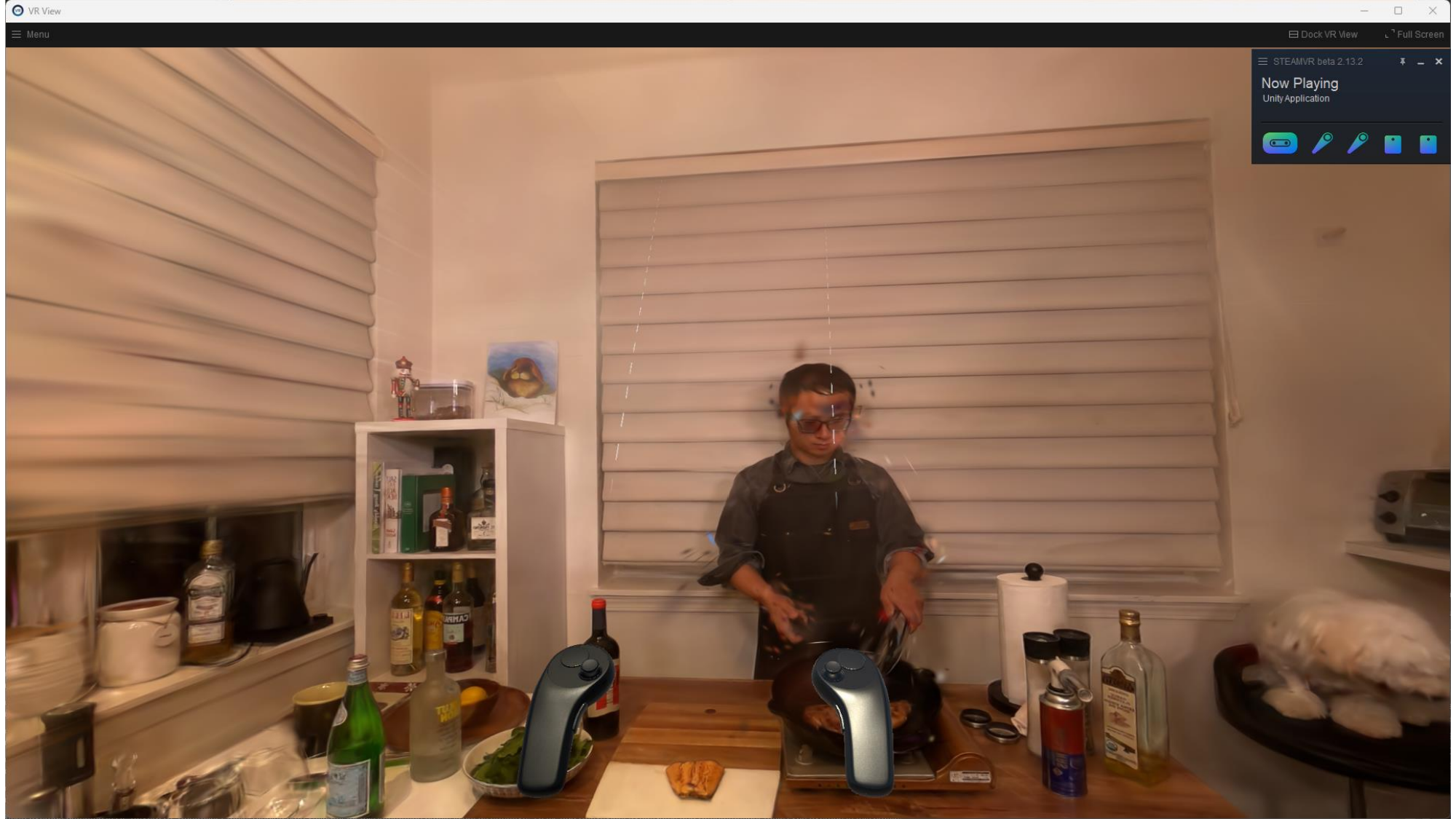}
  \caption{Real-time rendering performance on VR devices.}
  \label{fig:VR}
\end{figure}

By contrast, our improved representation substantially reduces model complexity and computational cost, making real-time rendering feasible under VR conditions. As shown in Fig.~\ref{fig:VR}, we develop a dedicated VR playback system that streams dynamic 3D Gaussian scenes directly into head-mounted displays (HMDs). This enables users to interactively explore dynamic scenes in immersive VR, moving beyond the limitations of 2D monitor demonstrations.

A discrepancy between the Unity and OpenGL coordinate systems leads to mirroring artifacts in our rendered scenes. Furthermore, while necessary optimizations are applied to adapt the 3DGS format for Unity rendering, these adjustments inevitably introduce aliasing, resulting in slightly lower visual fidelity compared to conventional 2D rendering.

\section{Limitations and Future Directions}
While HGS demonstrates substantial improvements, several open challenges remain as natural extensions of our work.

\textbf{Segmentation dependency.} Our method partially relies on the accuracy of static–dynamic separation. The precision of existing approaches is sufficient for our needs; nevertheless, we conducted additional tests to verify the robustness of our framework. We explicitly evaluated three possible error cases: minor boundary inaccuracies (mitigated by our two-stage training, Tab. ~\ref{tab:ablation}, Fig.~\ref{fig:ablation}), static regions mislabeled as dynamic (negligible impact, Tab.~\ref{tab:dyn-prop}), and dynamic regions mislabeled as static (0.8 dB PSNR drop, alleviated by our strategy with optional manual correction). These results suggest that HGS is robust to typical segmentation errors, and future work may integrate adaptive segmentation refinement.

\textbf{Temporal motion modeling.} As a local, kernel-based approach, explicit temporal RBF modeling may face challenges in capturing highly complex non-rigid motions. To address this, we employ a two-stage training strategy that allows dynamic Gaussians to appear or disappear over time, which proves effective in practice. Future extensions could combine RBF with more expressive formulations to better capture fine-grained deformations.

\textbf{Training strategy.} Our two-stage training design suppresses boundary artifacts and improves convergence stability. Although this introduces modest computational overhead, the trade-off is beneficial for quality, and exploring more streamlined training schedules will further enhance scalability to large-scale or ultra high-resolution scenarios.

\section{Conclusion}
We propose Hybrid Gaussian Splatting (HGS), a novel framework for dynamic scene reconstruction. By decomposing scenes into static and dynamic regions within a unified representation, HGS reduces parameter redundancy and improves computational efficiency. Our static and dynamic decomposition (SDD) strategy shares compact, temporally invariant parameters for static primitives, while a two-stage training strategy improves consistency at the boundaries between static and dynamic regions. Experiments demonstrate that HGS delivers comparable or superior rendering quality to state-of-the-art methods, while reducing model size by up to 98\% and achieving real-time 4K rendering at 125 FPS on an RTX 3090. It further sustains 160 FPS at $1352{\times}1014$ on an RTX 3050, and has been successfully integrated into the VR system, corroborating its practicality for immersive applications. Qualitative results further demonstrate that HGS preserves fine textures and mitigates artifacts caused by abrupt scene changes, setting a new benchmark for efficient, high-fidelity dynamic view synthesis.

\acknowledgments{
The authors wish to thank A, B, and C. This work was supported in part by
a grant from XYZ.}

\bibliographystyle{abbrv-doi}

\bibliography{template}
\end{document}